\def\year{2022}\relax
\documentclass[letterpaper]{article} 
\usepackage{aaai22}  
\usepackage{times}  
\usepackage{helvet}  
\usepackage{courier}  
\usepackage[hyphens]{url}  
\usepackage{graphicx} 
\usepackage{natbib}  
\usepackage{caption} 
\DeclareCaptionStyle{ruled}{labelfont=normalfont,labelsep=colon,strut=off} 
\usepackage{algorithm}
\usepackage{algorithmic}

\usepackage{paralist}
\usepackage{multirow}

\usepackage{newfloat}
\usepackage{listings}
\usepackage{todonotes}
\floatstyle{ruled}
\newfloat{listing}{tb}{lst}{}
\floatname{listing}{Listing}

\usepackage{CJKutf8}
\usepackage{amsmath}
\usepackage{xcolor}
\newcommand\Tstrut{\rule{0pt}{2.6ex}}         
\newcommand\Bstrut{\rule[-0.9ex]{0pt}{0pt}}   

\title{SexWEs: Domain-Aware Word Embeddings via Cross-lingual Semantic Specialisation for Chinese Sexism Detection in Social Media}

\author{
    Aiqi Jiang,
    Arkaitz Zubiaga
}
\affiliations{
    Queen Mary University of London\\

    \{a.jiang, a.zubiaga\}@qmul.ac.uk
}

\usepackage{bibentry}

\begin{document}
\begin{CJK*}{UTF8}{gbsn} 
\maketitle

\begin{abstract}

The goal of sexism detection is to mitigate negative online content targeting certain gender groups of people. However, the limited availability of labeled sexism-related datasets makes it problematic to identify online sexism for low-resource languages.
In this paper, we address the task of automatic sexism detection in social media for one low-resource language -- Chinese. Rather than collecting new sexism data or building cross-lingual transfer learning models, we develop a cross-lingual domain-aware semantic specialisation system in order to make the most of existing data. Semantic specialisation is a technique for retrofitting pre-trained distributional word vectors by integrating external linguistic knowledge (such as lexico-semantic relations) into the specialised feature space.
To do this, we leverage semantic resources for sexism from a high-resource language (English) to specialise pre-trained word vectors in the target language (Chinese) to inject domain knowledge. We demonstrate the benefit of our sexist word embeddings (\textsc{S}ex\textsc{We}s) specialised by our framework via intrinsic evaluation of word similarity and extrinsic evaluation of sexism detection. 
Compared with other specialisation approaches and Chinese baseline word vectors, our \textsc{S}ex\textsc{We}s shows an average score improvement of 0.033 and 0.064 in both intrinsic and extrinsic evaluations, respectively.
The ablative results and visualisation of \textsc{S}ex\textsc{We}s also prove the effectiveness of our framework on retrofitting word vectors in low-resource languages. 

\end{abstract}

\section{Introduction}


Due to the volume of incidents, hostile behaviours and violence in social media, manual inspection and moderation have become unmanageable, especially for minorities and minorised communities \citep{jha2017does,fersini2020ami,rodriguez2020automatic}. 
Sex is commonly a sensitive topic, and sexist content is of high subjectivity. The high cognition and tolerance thresholds of hostile gender-biased behaviour by certain gender groups can exacerbate gender-based hatred and violence online \citep{shi2020perception}.
Sexist speech refers to those promoting gender-based abuse and violence against an individual or a gender group of people on actual or perceived aspects of personal characteristics (e.g. physical gender differences) \citep{jiang2022swsr}, manifested in various behaviours (e.g. stereotyping, ideological issues, and sexual violence) \cite{manne2017down,anzovino2018automatic}.
\citet{glick2001ambivalent} define sexism as an ambivalent attitude manifested through both hostility and benevolence. Hostile sexism is characterised by an explicitly negative attitude towards gender groups (e.g. misogyny), while benevolent sexism is more subtle with seemingly positive characteristics.
Most studies focus more on detecting hostile sexism, overlooking implicit expressions of sexism \cite{waseem2016hateful,pamungkas2020misogyny}. Hence, mitigating online sexism in a wide spectrum of sexist behaviours is crucial as these are, in fact, extremely dangerous and harmful to society \cite{richardson2018woman}.


Research in sexism detection has recently increased in popularity \cite{rodriguez2020automatic,jiang2021exist}. However, sexism-related resources are predominantly available in English \citep{waseem2016hateful,jha2017does,samory2021callme} and Indo-European languages \citep{fersini2020ami,chiril2020annotated}, while efforts in low-resource languages are limited, such as Chinese \citep{jiang2022swsr}. To overcome this resource scarcity, cross-lingual transfer learning can be a solution.
Most studies in sexism detection focus on investigating the superior model architecture for classification in different languages
\citep{parikh2019multi,rodriguez2020automatic,samory2021callme} without using additional domain knowledge, such as a domain-specific lexicon \citep{wiegand2018inducing}.
Several works demonstrate the positive influence on the broader abusive language detection task
by directly injecting external domain knowledge at the model level \cite{koufakou2020hurtbert}, but lack further exploration into the effect of this knowledge.


Integrating structured external knowledge like distinct lexico-semantic relations into the feature space yields better performance in various downstream tasks, such as spoken language understanding \citep{kim2016adjusting}, text simplification \citep{ponti2018adversarial}, and cross-lingual transfer of resources \citep{vulic2017cross,ponti2019cross}. Semantic specialisation, referred to as retrofitting or post-processing, is a process of fine-tuning pre-trained distributional word vectors by incorporating structured linguistic constraints from external lexical resources (e.g., WordNet or BabelNet) to highlight specific semantic relations in the specialised embedding space, leading to the benefit of downstream applications \citep{faruqui2015retrofitting,mrksic2017semantic}. However, to overcome the restriction of constraint-driven specialisation only for existing (seen) words, a post-specialisation technique is proposed to leverage implicit information extracted from an initially specialised vector space to further specialise the entire vector space (on unseen words) \citep{vulic2018post,glavas2018explicit}. In addition, 
post-specialisation approaches are designed for cross-lingual scenarios by transferring global specialisation via a shared vector space \citep{glavas2019properly,ponti2019cross}. Previous studies have developed semantic specialisation techniques for distributional word embeddings \citep{mrksic2017semantic} and contextualised embeddings with sentence-level semantics \citep{vulic2021lexfit}. As far as we know, no previous work has studied the fine-tuning of word embeddings with domain-specific semantic knowledge through cross-lingual semantic specialisation techniques for a low-resource social media task such as sexism detection.

In this paper, we develop a domain-aware cross-lingual semantic specialisation framework between languages (i.e. English-to-Chinese), aiming to construct sexism-specific word embeddings (\textsc{S}ex\textsc{We}s) to facilitate the performance of the sexism detection task for a low-resource language. 
Inspired by the cross-lingual specialisation method in \citet{ponti2019cross}'s work, in our case, we first structure linguistic constraints from external sexism-related semantic knowledge (e.g. BabelNet) into different forms, including source constraints (English), target constraints (Chinese) and cross-lingual constraints. Then we project all source constraints into target constraints, and refine these projected target constraints by cleaning up the noise inside them. After that, various target constraint groups are incorporated together into the specialisation process to retrofit pre-trained word embeddings to be domain-aware for the target language. Finally, we can monolingually employ our domain-specific \textsc{S}ex\textsc{We}s to the downstream task of social media sexism detection. 

Furthermore, we verify the quality of our \textsc{S}ex\textsc{We}s in the intrinsic evaluation of word similarity, as well as the impact on sexism detection. Our results show that \textsc{S}ex\textsc{We}s achieves state-of-the-art performance on several word similarity benchmarks, outperforming all baseline classifiers on identifying sexism. Additionally, the visualisation of \textsc{S}ex\textsc{We}s with diverse constraints shows positive changes before and after the specialisation, and an ablation study also demonstrates the effectiveness of our proposed architecture for cross-lingual domain-aware specialisation. Our specialisation method enables us to specialise any type of distributional vectors in the target language with diverse constraints.

Our key contributions include the following:
\begin{enumerate}
    \item We conduct the first study on semantic specialisation for cross-lingual abusive language detection, building \textsc{S}ex\textsc{We}s, sexism-specific word embeddings for Chinese;
    \item Our domain-aware embeddings achieve state-of-the-art performance on word similarity benchmarks (correlation score increased by 0.039) and the Chinese sexism detection task compared with all Chinese baseline embeddings (F1 score improved by 0.114);
    \item Our cross-lingual domain-aware specialisation outperforms previous state-of-the-art specialisation transfer approaches on both word similarity benchmarks (correlation score increased by 0.027) and the Chinese sexism detection task (F1 score improved by 0.041);
    \item We will publicly release our resources\footnote{https://github.com/aggiejiang/SexWEs} to facilitate the integration of external lexical domain knowledge into distributional embedding models for other low-resource languages.
\end{enumerate}

\section{Related Work}

\subsection{Sexism Detection}


Research in social media sexism detection has increased in recent years \citep{parikh2019multi,samory2021callme}. The first attempt was by \citet{hewitt2016problem} who investigated the manual classification of gender-based tweets, and the first survey of automatic misogyny identification in social media was conducted by \citet{anzovino2018automatic}. \citet{rodriguez2020automatic} explore the feasibility of automatically identifying sexist content using both traditional and deep learning techniques.
In addition, researchers mainly address the problem of multilingual sexism detection by using deep neural networks with cross-Lingual word embeddings or multilingual pre-trained models \citep{pamungkas2020misogyny,rodriguez2021overview}.
However, most relevant studies investigate monolingual or multilingual sexism detection only based on existing data in high-resource languages such as English and other Indo-European languages \citep{fersini2020ami,chiril2020annotated,samory2021callme}, while cross-lingual studies in the field of sexism and even general abuse are still limited for low-resource languages like Chinese.


Moreover, most studies propose model architectures for identifying online sexism or related abuse, and few make efforts to infuse external domain knowledge into vector space to enhance the detection performance \citep{arango2021cross}. \citet{badjatiya2017deep} utilise an \textsc{Lstm}-based model to generate English hate word embeddings, but more persuasive validation strategies should be reconsidered \citep{arango2020hate}. \citet{kamble2018hate} describe the construction of domain word embeddings based on Word2Vec from a Hindi-English hate speech dataset, and \citet{alatawi2021detecting} produce abuse-specific embeddings for English white supremacy. Besides, multilingual word embeddings based on abuse knowledge are created for cross-lingual hate speech detection \citep{arango2021cross}. 


To the best of our knowledge, no prior work has studied cross-lingual semantic specialisation techniques to generate domain-aware word embeddings in low-resource languages for abuse in social media. Therefore, we specifically apply this technique to the field of sexism and choose Chinese\footnote{Chinese is generally a resourceful language, but there is only one dataset available for online abuse or sexism.} as our target language. Abusive language detection is in turn understudied in Chinese, with the only antecedent of \citep{jiang2022swsr}, who created the sexism dataset that we use here but didn't study model development.

\subsection{Retrofitting Word Embeddings}


In the field of word vector specialisation, there has been a body of research exploring various methods to incorporate diverse constraints into the word embedding space. The first retrofitting work by \citet{faruqui2015retrofitting} is proposed to pull the vectors of similar words closer to each other by fusing only synonyms. Then \textsc{Attract}-\textsc{Repel}, a standard semantic specialisation approach, is developed to integrate structured linguistic constraints with both similar and dissimilar semantics into pre-trained vector spaces, clustering the embeddings of similar words (e.g. synonyms, hypernym-hyponym pairs) closer together and enforcing dissimilar words (e.g., antonyms) far away from each other \citep{mrksic2017semantic}. Such semantic specialisation could be applied to any kind of distributional word embeddings.

Since the first-generation semantic specialisation models only retrofit the embeddings of words seen in linguistic constraints, a series of post-specialisation techniques are proposed \citep{vulic2018post,glavas2018explicit,ponti2018adversarial,colon-hernandez2021retrogan}. Post-specialisation aims to fine-tune the entire distributional vector space by learning an explicit and global specialisation mapping between original and initially specialised spaces, and then applying the mapping to the embeddings of words unseen in external constraints \citep{vulic2018post}. \citet{ponti2018adversarial,colon-hernandez2021retrogan} modify the feed-forward post-specialisation network with different Generative Adversarial Networks (\textsc{Gan}s) based approaches to discriminate word vectors from original and specialised spaces, which yields better performance on retrofitting. 

Post-specialisation approaches can be further employed for cross-lingual transfer through a shared vector space between source and target languages \citep{glavas2019properly,ponti2019cross}. In this work, we demonstrate how to combine task-oriented multilingual domain knowledge to achieve cross-lingual semantic specialisation on pre-trained word embeddings, with an impact on sexism detection for low-resource languages.

\section{Methodology: \textsc{S}ex\textsc{We}s}

\begin{figure*}[h]
  \centering
  \includegraphics[width=0.99\linewidth]{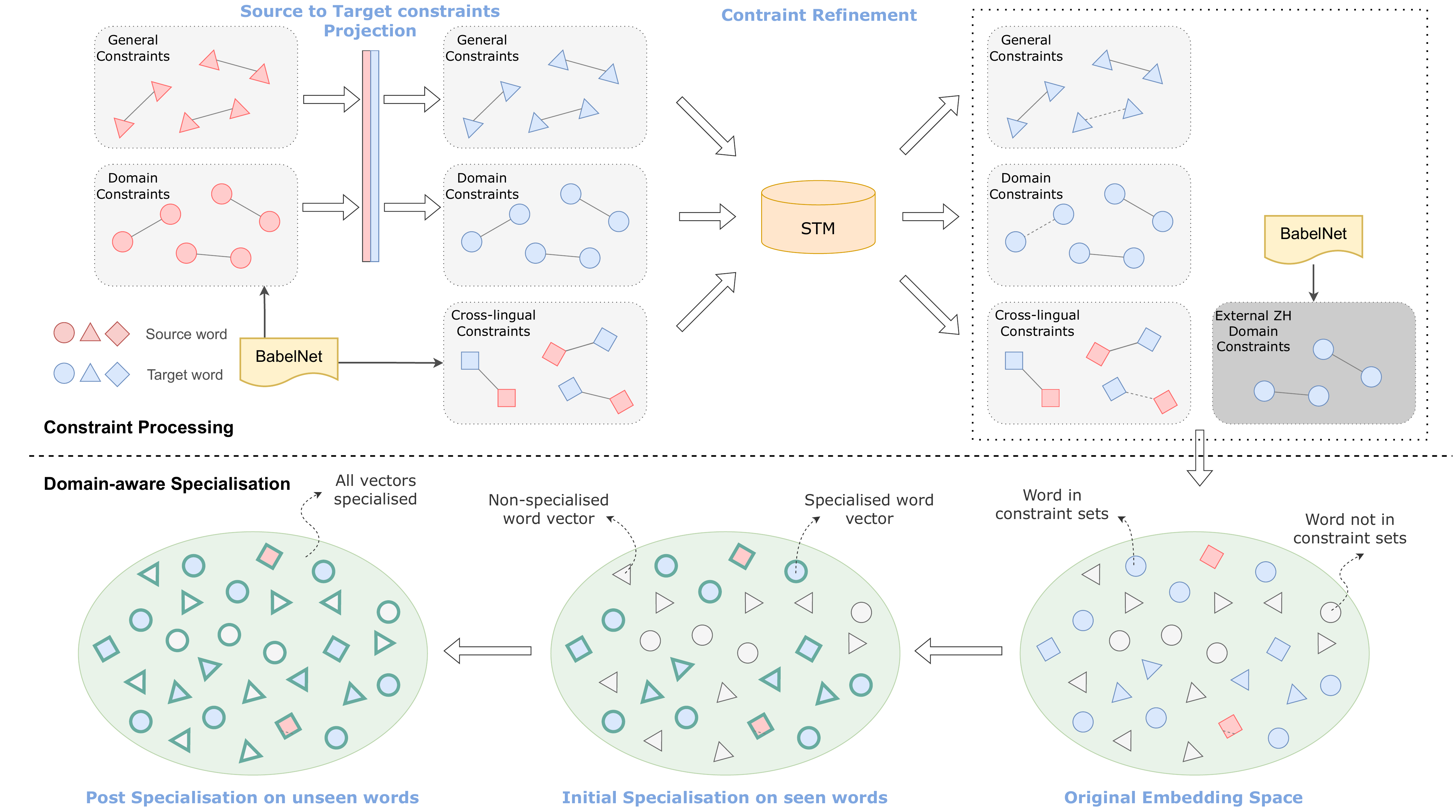}
  \caption{Overview of \textsc{S}ex\textsc{We}s. Constraint processing collects multilingual domain constraints, projects them across languages and filters noisy pairs. Domain-aware specialisation retrofits distributional word vectors in two steps: (1) utilise knowledge-aware constraints to specialise vectors on seen words; (2) learn and apply specialised mapping to the entire space.}
  \label{fig:model}
\end{figure*}

We propose to build Sexist Word Embeddings (\textsc{S}ex\textsc{We}s) based on a cross-lingual domain-aware semantic specialisation system, inspired by the \textsc{Clsri}  framework \citet{ponti2019cross}. The objective is to incorporate awareness of the sexism domain into the semantic specialisation procedure to enrich domain-aware word embeddings (integrated sexism domain knowledge). We aim to specialise existing state-of-the-art distributional word embeddings in a target language by utilising commonsense knowledge and multilingual domain knowledge from lexical constraints, where constraints are dominated by resource-rich source language and supplemented by a resource-poor target language. In our case, we opt for English (\textsc{En}) as the source language $L_{en}$ and Chinese (\textsc{Zh}) as the target language $L_{zh}$. 

Our procedure can be split into two parts: constraint processing and domain-aware specialisation (see Figure \ref{fig:model}). Firstly, constraint processing is to collect multilingual domain constraints, project source constraints across languages and clean up noisy constraints by transformation. Then we fuse the refined target constraints and external target constraints together as constraints $C^{group}_{zh}$, and execute monolingually initial specialisation and post-specialisation on existing distributional word vector space by employing well-handled constraints $C^{group}_{zh}$ in the target language.

\subsection{Constraint Processing}

According to \citet{mrksic2017semantic}'s \textsc{Attract}-\textsc{Repel} methodology, linguistic constraints obtained from external sources are usually divided into two lexico-semantic groups: 

\begin{itemize}
    \item \textit{\textsc{Attract} constraints}: indicate word pairs with similar representations, e.g. synonyms (swearing and abuse, 咒骂 and 辱骂) or direct hypernym-hyponym pairs (woman and widow, 女人 and 寡妇); 
    \item \textit{\textsc{Repel} constraints}: specify which word pairs should appear far apart in the vector space, e.g. antonyms (appreciation and disgust, 欣赏 and 厌恶).
\end{itemize}

Our constraints are grouped into five categories:

\begin{itemize}
    \item English general constraints $C_{en}^{g}$ 
    \item English domain constraints $C_{en}^{d}$
    \item English general\&domain constraints $C_{en}^{both} = C_{en}^{g} \cup C_{en}^{d}$
    \item Chinese domain constraints $C_{zh}^{d}$
    \item Cross-lingual \textsc{En}-\textsc{Zh} domain constraints $C_{cl}^{d}$
\end{itemize}

English general constraints include words that are commonly and frequently used, while domain constraints refer to words related to the domain. 
In our case, we continue to use the existing general constraints \citep{ponti2019cross} and extract domain constraints in both monolingual and cross-lingual scenarios.
Except for $C_{cl}^{d}$ constraints, the other four types of constraints all have \textsc{Attract} and \textsc{Repel} sets separately. This step focuses on processing source constraints and cross-lingual constraints while target constraints $C_{zh}^{g}$, $C_{zh}^{d}$ or $C_{zh}^{both}$ could be regarded as external constraints to facilitate specialisation performance in the next step.

In this constraint processing step, we first collect domain constraints into \textsc{Attract} and \textsc{Repel} sets from BabelNet in source language $C_{en}^{d}$ and target language $C_{zh}^{d}$, and project all constraints in source language $C_{en}^{d}$ to those in target language $C^{d}_{zh^{'}}$. Considering imperfect mapping and polysemy of $C_{en}^{d}$ possibly leading to the incorrect meaning of $C^{d}_{zh^{'}}$, these noisy constraints $C^{d}_{zh^{'}}$ are filtered via a variant of Specialisation Tensor Model (\textsc{Stm}) \citep{glavas2018discriminating}.

\subsubsection{Monolingual and Cross-lingual Domain Constraints Collection}

In order to extract monolingual domain constraints, we organise domain seed words from several domain-related lexical resources for both source $C_{en}^{d}$ and target $C_{zh}^{d}$ languages separately. Then we create domain constraint pairs via searching synonyms and antonyms in the same language for each seed word, and add a language tag before each word, such as (zh\_歧视, zh\_偏见)\footnote{歧视 or 偏见 means an unfair and unreasonable opinion or feeling, especially when formed without enough thought or knowledge, such as prejudice or bias.}

In addition to monolingual domain constraints, we also extract cross-lingual domain constraints $C_{cl}^{d}$ based on domain seed words in the form of English-Chinese constraints. It will be taken into consideration such as explicit and implicit cross-lingual domain constraints. An explicit constraint refers to those English-Chinese constraints via direct translation and both words are explicitly domain-related (such as (en\_prejudice, zh\_歧视)), while an implicit constraint means two words cannot be directly translated from/to each other, because one word is domain-related in one language but another one could be domain-unrelated in another language if directly translated (such as (en\_f*cking, zh\_草)\footnote{The primary meaning of 草 is grass, only in certain occasions it may mean the same as f*cking.} and (zh\_绿茶婊, en\_angelic b*tch)\footnote{绿茶婊 refers to girls who pretend to be pure and innocent but in fact are manipulative and scheming. It literally translates into green tea b*tch. The meaning of 绿茶婊 is similar to angelic b*tch}).
All domain seed words are first directly translated\footnote{We use Google Translate https://translate.google.co.uk/.} into explicit constraints, then we manually check and correct incorrectly translated word pairs to generate implicit constraints.

\subsubsection{Source to Target Constraints Projection}

Learning cross-lingual word embeddings via supervised approaches shows good performance on the task of Bilingual Lexicon Induction (BLI) especially on typologically-distant language pairs like \textsc{En}-\textsc{Zh} \citep{wang2021multi}. Recent work \citep{wang2021multi} has shown that Relaxed Cross-domain Similarity Local Scaling (RCSLS) \citep{joulin2018loss}, as a supervised system, achieves remarkable performance among competing models on the BLI task, and it has been applied to the word translation task in order to enhance the performance. So we leverage the RCSLS model to learn a linear cross-lingual projection matrix $\textbf{W}_{en\_zh}$ between source and target word embeddings.

Given a set of source constraints $C_{en}$, each constraint is presented as a word pair $(w_{en}^{a}, w_{en}^{b})$. Since phrases exist widely in domain constraints $C_{en}^{d}$, phrase-level projection is also employed by averaging all word embeddings per phrase.
We translate each word or phrase $w_{en}$ in source constraints by looking for the nearest neighbour of its (averaged) embedding $\textbf{x}_s$ in the projected target space. 
We project source constraints $C_{en}$ into target constraints $C^{d}_{zh^{'}}$ by using the projection matrix $\textbf{W}_{en\_zh}$ to project source and target embeddings into a shared bilingual space $\textbf{X}_{en\_zh}$.


\subsubsection{Target Constraint Refinement}

The shared bilingual space obtained by the cross-lingual projection matrix is far from perfect due to incorrect translation via the cross-lingual shared space and incorrect of senses of polysemous words in $L_{en}$.
Hence, noisy constraints could be generated via projection-based approaches from source constraints $C_{en}$ to target constraints $C^{d}_{zh^{'}}$ \citep{glavas2019properly}.

Similar to the \textsc{Clsri}  framework, we aim to purify noisy constraints in $C^{d}_{zh^{'}}$ by leveraging the Specialisation Tensor Model (\textsc{Stm}) to discriminate lexico-semantic relations within word pairs \citep{glavas2018discriminating}. \textsc{Stm} is a simple and effective feed-forward neural architecture that predicts lexical relations between word pairs by specialising input distributional word embeddings in multiple different projections and computing latent scores from these specialisation tensors for the final relation classifier. \textsc{Stm} performs better particularly for synonyms and antonyms, and also presents stable performance across languages \citep{glavas2018discriminating}. We alter the multi-label \textsc{Stm} classifier to a binary classifier\footnote{See more \textsc{Stm} technical details in \citet{glavas2018discriminating}}, and train five types of instances for \textsc{Stm}:

\begin{itemize}
    \item G$_a$-\textsc{Stm} \& D$_a$-\textsc{Stm}: it predicts whether a word pair from general or domain constraints represents a valid \textsc{Attract} constraint;
    \item G$_r$-\textsc{Stm} \& D$_r$-\textsc{Stm}: it predicts whether a word pair from general or domain constraints represents a valid \textsc{Repel} constraint; 
    \item D$_{cl}$-\textsc{Stm}: it predicts whether a pair of cross-lingual domain words represents a valid \textsc{Attract} constraint;
\end{itemize}


\subsection{Domain-Aware Specialisation}
\label{sec:specialisation}

The step of domain-aware specialisation consists of monolingually retrofitting distributional word embeddings space in the target language $L_{zh}$ by leveraging a group of target constraints，such as projected target constraints (e.g. $C^{g}_{zh^{'}}$, $C^{d}_{zh^{'}}$ or $C^{both}_{zh^{'}}$) plus external target constraints (e.g. $C_{zh}^{g}$, $C_{zh}^{d}$ or $C_{zh}^{both}$). The whole semantic specialisation process is similar to the \textsc{Clsri}  system \citep{ponti2019cross}. 
Following the state-of-the-art specialisation model \textsc{Attract}-\textsc{Repel} (AR) \citep{mrksic2017semantic},
we initially specialise the target distributional space to be domain-aware but limited to existing $C_{zh}$ constraints. Then, based on the \textsc{Ar} specialisation, we apply the state-of-the-art post-specialisation model \textsc{Retrogan} \citep{colon-hernandez2021retrogan} to the entire vocabulary $V_{zh}$, including all the words seen and unseen in the target space. The following is a detailed description of system and a brief outline of \textsc{Ar} and \textsc{Retrogan} models.

\subsubsection{Initial Domain-Aware Specialisation}

The group of target constraints $C^{group}_{zh}$ to be specialised is a combination of projected target constraints $C_{zh^{'}}$ from source constraints $C_{en}$ and external target constraints
$C_{zh}$ from scratch, where $C^{group}_{zh} = C^{type}_{zh} \cup C^{type}_{zh^{'}}$ and $type = \{g,d,both\}$. After the combination, $C^{group}_{zh}$ includes two constraint subsets: \textsc{Attract} constraints $A_{zh}$ and \textsc{Repel} constraints $R_{zh}$. The distance of each word pair $(w_{zh}^{a}, w_{zh}^{b})$ from $A_{zh}$ and $R_{zh}$ is refined between their corresponding embeddings $(\textbf{x}_{zh}^{a}, \textbf{x}_{zh}^{b})$ in the target distributional space.

The specialisation process is carried out via mini-batches of $C^{group}_{zh}$.
Let $\mathcal{B}_A$ be a batch of vector pairs from $A_{zh}$ and $\mathcal{B}_R$ the batch from $R_{zh}$. We define $\mathcal{T}_A(\mathcal{B}_A)$ and $\mathcal{T}_R(\mathcal{B}_R)$ as corresponding negative pairs for each $\mathcal{B}_A$ and $\mathcal{B}_R$. For each $A_{zh}$ (or $R_{zh}$) constraint $(\textbf{x}_{zh}^{a}, \textbf{x}_{zh}^{b})$, we retrieve its closest (or farthest) vector pair as the negative constraint $(\textbf{t}_{zh}^{a}, \textbf{t}_{zh}^{b})$. Half of the negative constraints are selected based on their cosine similarity, and the other half are random negative samples.

The objective of \textsc{Ar} retrofitting is to minimise the max margin loss between target constraints and their corresponding negative samples, which includes three types of losses: 
\begin{equation}
    \mathcal{L}_{AR} = Att(\mathcal{B}_A,\mathcal{T}_A) + Rep(\mathcal{B}_R,\mathcal{T}_R) + Pre(\mathcal{B}_A, \mathcal{B}_R)
\end{equation}
Specifically, $Att(\mathcal{B}_A,\mathcal{T}_A)$ enables target constraints in $\mathcal{B}_A$ closer together than those in the corresponding $\mathcal{T}_A$ by a \textsc{Attract} margin $\delta_A$:
\begin{equation}
\label{eq:attract}
\begin{split}
    Att(\mathcal{B}_A,\mathcal{T}_A) = & \sum_{i=1}^{|\mathcal{B}_A|}[\mathcal{T}(\delta_A+\textbf{x}_{zh_i}^{a}\textbf{t}_{zh_i}^{a}-\textbf{x}_{zh_i}^{a}\textbf{x}_{zh_i}^{b}) \\ & +\mathcal{T}(\delta_A+\textbf{x}_{zh_i}^{b}\textbf{t}_{zh_i}^{b}-\textbf{x}_{zh_i}^{a}\textbf{x}_{zh_i}^{b})]
\end{split}
\end{equation}
where $\mathcal{T}(x) = max(0, x)$ is the hinge loss function, and $\delta_A$ determines how much closer target constraints from $A_{zh}$ are to each other than the distance to their corresponding negative examples. Analogously, $Rep(\mathcal{B}_R,\mathcal{T}_R)$ imposes constraints in $\mathcal{B}_R$ farther than their corresponding constraints in $\mathcal{T}_R$ based on a \textsc{Repel} margin $\delta_R$. Besides, $Pre(\mathcal{B}_A, \mathcal{B}_R)$ is the regularisation term to preserve the high-quality semantic information from $\textbf{X}_{zh}$ by minimising the Euclidean distance between plain and specialised embeddings. 

After \textsc{Ar} specialisation, \textsc{Ar} specialised space $\textbf{X}^{'}_{zh} \in \mathcal{R}^d$ is generated from the initial distributional space $\textbf{X}_{zh} \in \mathcal{R}^d$.

\subsubsection{Cyclic Adversarial Post-Specialisation} 

The \textsc{Ar} specialisation only works on the target words $V^{seen}_{zh}$ that actually exist in $C^{group}_{zh}$, which indicates that the performance of initial specialisation can be semantically improved in terms of the overlapping vocabulary between explicit $V^{seen}_{zh}$ and the vocabulary $V_{zh}$ of the initial distributional space $\textbf{X}_{zh}$. Post-specialisation learns the mapping from initial specialisation space and propagates it to the rest of the vocabulary $V^{unseen}_{zh}$ \citep{vulic2018post,colon-hernandez2021retrogan}.

\textsc{Retrogan} enriches the existing adversarial post-specialisation model \citep{ponti2018adversarial} to a Cycle\textsc{Gan}-like architecture with a pair of Generative Adversarial Networks (\textsc{Gan}s) \citep{goodfellow2014generative}. The goal of \textsc{Retrogan} is to learn a global specialisation mapping by balancing a combination of losses in both post-specialisation and inversion to ensure a unique one-to-one mapping between the plain vector space $\textbf{X}_{zh}$ and specialised \textsc{Ar} space $\textbf{X}^{'}_{zh}$ as conditioned by embeddings of seen words $V^{seen}_{zh}$ from $C^{group}_{zh}$ constraints. Then it propagates this global mapping to the entire distributional space of our target language $\textbf{X}_{zh}$.

The model combines both cyclic and non-cyclic optimisation objectives, where the contrastive margin-based ranking loss with random confounders $L_{MM}$ \citep{ponti2018adversarial,ponti2019cross} is used for both the generators and additionally for the cycle of generators\footnote{See more technical details of the \textsc{Retrogan} and its losses in \citet{colon-hernandez2021retrogan}}:
\begin{equation}
\label{eq:retrogan_loss}
\begin{split}
    & \mathcal{L}_{MM} = \sum_{i=1}^{|| V^{seen}_{zh}||}\sum_{j=1|j\neq i}^{k}  \mathcal{T}\\ & [(\delta_{MM}-\cos(G(\textbf{x}_{zh_{i}}),\textbf{x}^{'}_{zh_{i}})+\cos(G(\textbf{x}_{zh_{i}}),\textbf{x}^{'}_{zh_{j}}) + \\ &
    (\delta_{MM}-\cos(F(\textbf{x}_{zh_{i}}),\textbf{x}^{'}_{zh_{i}})+\cos(F(\textbf{x}_{zh_{i}}),\textbf{x}^{'}_{zh_{j}}) + \\&
    (\delta_{MM}-\cos(G(F(\textbf{x}_{zh_{i}})),\textbf{x}^{'}_{zh_{i}})+\cos(G(F(\textbf{x}_{zh_{i}})),\textbf{x}^{'}_{zh_{j}}) + \\&
    (\delta_{MM}-\cos(F(G(\textbf{x}_{zh_{i}})),\textbf{x}^{'}_{zh_{i}})+\cos(F(G(\textbf{x}_{zh_{i}}),\textbf{x}^{'}_{zh_{j}}))]
\end{split}
\end{equation}
where $G: \textbf{X}_{zh} \rightarrow \textbf{X}^{'}_{zh}$ is the generator that maps the plain vector space $\textbf{X}_{zh}$ to the specialised space $\textbf{X}^{'}_{zh}$, and $F: \textbf{X}^{'}_{zh} \rightarrow \textbf{X}_{zh}$ is the generator that does the opposite. $L_{MM}$ makes a word vector generated from $\textbf{X}_{zh}$ by generators closer to its gold-standard vector (e.g. specialised \textsc{Ar} vector $\textbf{x}^{'}_{zh} \in \textbf{X}^{'}$) and different from any of $k$ random confounders by a margin $\delta_{MM}$, and then forces this constraint across the cycle.

\section{Experimental Setup}

\subsection{Initial Distributional Word Embeddings}

As a starting point to build domain-aware specialised embeddings, we employ publicly available \textsc{Fasttext} word vectors \citep{grave2018learning} for both English and Chinese.\footnote{Other multilingual embedding models, such as LASER, Multilingual \textsc{Bert} and XLM-R, could be tested, however they are generally better suited for sentence-level embeddings.} They provide 300-dimensional word vectors trained on Common Crawl and Wikipedia in 157 languages, using CBOW with position weights. We execute the projection from source to target vector space via supervised RCSLS method, searching 10 nearest neighbours in 10 iterations.

\subsection{External Sexism Lexical Knowledge}

To generate domain-specific constraints, we intend to use some lexical resources related to sexist domains to organise sexist seed words. However, due to the lack of external resources specifically addressing sexism, we select words from abusive language-related resources, where abuse is a superdomain of sexism \citep{waseem2016hateful}.

For the source language (\textsc{En}), we use (i) the hate speech lexicon \textit{HurtLex}, containing 6,287 seed offensive, aggressive, and hateful words and phrases in over 50 languages \citep{bassignana2018hurtlex},
and (ii) the abuse lexicon by \textit{\citet{wiegand2018inducing}}, which includes 2,989 words.
For the target language (\textsc{Zh}), we use \textit{SexHateLex} \citep{jiang2022swsr}, a large Chinese sexism lexicon including 3,016 profane and sexually abusive and slang words and phrases.

\subsection{Linguistic Constraints}

Linguistic Constraints are present in the form of word/phrase pairs in the source language (\textsc{En}) and the target language (\textsc{Zh}) for semantic specialisation, which is divided into three categories: source general constraints, bilingual domain (sexism) constraints and cross-lingual constraints. We also combine general and domain constraints (in the same language) as another group of constraints. The number of constraints is summarised in Table \ref{tab:constraints}.

\begin{itemize}
    \item \textit{Source General Constraints:} We follow the same English general constraints as used in previous work for the specialisation process \citep{
    ponti2018adversarial,ponti2019cross}. These general constraints involve the lexico-semantic relations from WordNet \citep{miller1995wordnet}, Paraphrase Database (PPDB) \citep{ganitkevitch2013ppdb} and BabelNet \citep{navigli2010babelnet}, which covers 16.7\% of the 200K most frequent English words in the vocabulary of \textsc{Fasttext} embeddings.
    \item \textit{Bilingual Domain Constraints:} To produce domain constraints, we employ the multilingual semantic network BabelNet
    on sexism-related seed words or phrases to extract synonyms and antonyms according to word sense tags. These constraints cover only 14.4\% and 4.2\% of the English and Chinese vocabulary from \textsc{Fasttext}.
    \item \textit{Cross-lingual Domain Constraints:} Cross-lingual sexism-related (domain) constraints are English-Chinese pairs extracted via multilingual BabelNet 
    based on domain seed words or phrases (e.g. en\_hate, zh\_憎恶).
\end{itemize}

\begin{table}[htb]
\centering
\caption{Collection of \textsc{Attract} and \textsc{Repel} constraints for source (\textsc{En}) and target (\textsc{Zh}). Both is the aggregate and deduplicated set of general and sexism-related constraints.}
\label{tab:constraints}
\resizebox{0.4\textwidth}{!}{%
\begin{tabular}{llccc}
\hline
   &      & \textbf{General} & \textbf{Sexism} & \textbf{Both} \Tstrut\Bstrut \\ \hline
English & \textsc{Attract} & 640,435           & 130,445        & 768,294       \Tstrut \\ 
    & \textsc{Repel} & 11,939            & 501           & 12,148     \Bstrut   \\ \hline
Chinese & \textsc{Attract} & -       & 6,353          & -  \Tstrut  \\ 
           & \textsc{Repel} & -       & 32            & -        \Bstrut   \\ \hline
\textsc{En}-\textsc{Zh} & \textsc{Attract} & -        & 189          & -      \Tstrut\Bstrut   \\ \hline
\end{tabular}%
}
\end{table}

\vspace{-0.1cm}
\subsection{Specialisation Approaches in Comparison}

We compare our \textsc{S}ex\textsc{We}s specialisation on different types of constraints with three other semantic specialisation methods, implemented using the same \textsc{Fasttext} embeddings and both constraints used for our model \textsc{S}ex\textsc{We}s:

\begin{itemize}
    \item \textit{\textsc{Attract}-\textsc{Repel} (\textsc{Ar}):} A state-of-the-art retrofitting approach \citep{mrksic2017semantic} to refine a distributional vector space by using \textsc{Attract}/synonymy and \textsc{Repel}/antonymy constraints. 
    \item \textsc{Retrogan} : A post-specialisation approach \citep{colon-hernandez2021retrogan} by learning the mapping of \textsc{Ar} and then extending an adversarial post-specialisation model Aux\textsc{Gan}  \citep{ponti2018adversarial} into a Cycle\textsc{Gan}-like architecture \citep{zhu2017cyclegan} on the entire dataset.
    \item \textit{\textsc{Clsri} :} A specialisation Transfer via Lexical Relation Induction \citep{ponti2019cross} transfers specialisation mapping from a resource-rich source language (English) to virtually any target language based on \textsc{Ar} and Aux\textsc{Gan} with noisy constraints cleanup.
\end{itemize}

\subsection{Hyperparameters in the Training Process}

\subsubsection{Constraints Refinement: \textsc{Stm}}

The \textsc{Stm} model is adopted to predict lexical relations between constraints with 5 specialisation tensors, 300 neurons of the hidden layer and a 0.5 dropout value based on prior work \citep{ponti2019cross}. During training, we set the batch size to 32 and the maximum number of iterations to 10, using Adam optimiser \citep{kingma2015adam} with a learning rate of 0.0001.

\subsubsection{Initial specialisation: \textsc{Ar}}

We preserve the hyperparameter settings for \textsc{Ar} as used by \citet{mrksic2017semantic}. The margins for \textsc{Attract}, \textsc{Repel} and regularisation are 0.6, 0.0 and $1e^{-9}$, respectively. The Adagrad optimiser \citep{duchi2011adagrad} is used with 0.05 learning rate, batch size is 50, and maximum number of iterations is 5. The same configuration as the baseline AR.

\subsubsection{Post-Specialisation: \textsc{Retrogan} }

We use two hidden layers with 2,048 units for the generator and the discriminator in each \textsc{Gan}  of \textsc{Retrogan} , adopting 0.2 and 0.3 dropout rates separately. We set the margin $\delta_{MM}$ to 1.0 and the number of negative samples to 25, utilising Adam optimiser with 0.1 learning rate. The number of training epochs is set to 10 and batch size 32, same as the baseline \textsc{Retrogan}  model.

\section{Results and Analysis}
\label{sec:result}

We evaluate our \textsc{S}ex\textsc{We}s via both intrinsic evaluation of word similarity and extrinsic evaluation of sexism detection.

\subsection{Intrinsic Evaluation : Word Similarity}

The first experiment is to assess the quality of our specialised space of \textsc{S}ex\textsc{We}s via the word similarity task, which aims to evaluate the ability of the model to capture the semantic proximity and relatedness between two words.

\subsubsection{Chinese Embeddings in Comparison}

We adopt original \textsc{Fasttext} word vectors and retrofitted vectors by other specialisation approaches in comparison with our specialised embeddings infusing diverse constraints.

\subsubsection{Evaluation Setup}

We employ three word similarity benchmarks, namely SimLex-999 (SL999) \citep{hill2015simlex}, WordSim-296 (WS296) \citep{peng2012semeval} and WordSim-240 (WS240) \citep{wang2011computing}. WS296 and WS240 are Chinese datasets, while SL999 is an English dataset then translated into traditional Chinese by \citet{su2017learning}. We convert it from traditional to simplified Chinese with chinese-converter\footnote{https://pypi.org/project/chinese-converter/}. The word pair coverage in the datasets is 975 of 999 for SL999, 230 of 240 for WS240, and 286 of 297 for WS296. 
The Spearman's rank correlation $\rho$ is measured as the intrinsic evaluation metric between the gold word pair similarity scores by annotators and the cosine similarity scores of the corresponding word embeddings from various vector spaces.

\subsubsection{Analysis of Results}

The results of word similarity tests are summarised in Table \ref{tab:similarity}. Regardless of whether we plus external Chinese domain constraints or not, our specialised \textsc{S}ex\textsc{We}s basically outperforms the initial distributional vectors (0.039) and other cross-lingual specialisation models (0.027), indicating the effectiveness of incorporating domain constraints in source language during the cross-lingual transfer. 
And to the best of our knowledge, our results also surpass the Chinese word embeddings \textsc{Vcwe} \citep{sun2019vcwe} that achieves the state-of-the-art performances on WS240 and WS296\footnote{The \textsc{Vcwe} results are 0.578 for WS240 and 0.613 for WS296, and it exceeds many competitive Chinese embeddings \citep{xiong2021learning}. For more results, see https://chinesenlp.xyz/docs/word\_embedding.html.}. By fusing external domain-specific target pairs, it also achieves better results for vector space specialisation.
Moreover, even without the infusion of sexist-related knowledge, our approach still outperforms the similarly structured model \textsc{Clsri}，while noticeably exceeding two separate models of \textsc{Ar} and \textsc{Retrogan}, respectively.
Although our \textsc{S}ex\textsc{We}s achieves a satisfactory performance on SL999 among all models, it can still be noted that there is no big gap compared to scores on the other two benchmarks, probably due to the translation issue from English to Chinese version or the conversion issue between traditional and simplified Chinese.

\begin{table}[h]
\centering
\caption{Results of word similarity evaluation based on Spearman's rank correlation score $\rho$ (average of 5 runs).} 
\label{tab:similarity}
\resizebox{0.33\textwidth}{!}{%
\begin{tabular}{lccc}
\hline
\textbf{}                              & \textbf{SL999}                                           & \textbf{WS240}                                          & \textbf{WS296}                                        \Tstrut\Bstrut  \\ \hline
\textsc{Fasttext}          &   .347           &   .546           &   .620        \Tstrut   \\
\textsc{Ar}            &   \underline{.402}    &   .521         &   .586        \\
\textsc{Retrogan}           &   .380          &   .572          &   .615          \\
\textsc{Clsri}              &   .384          &   .558         &   \underline{.627}  \Bstrut \\ \hline
\textbf{\textsc{S}ex\textsc{We}s} &   \textbf{.406} &   \textbf{.586} &   .608      \Tstrut   \\
w/o external   &   .394          &   \underline{.581}    &   .624         \\ only general   &   .389          &   .561          &   .623          \\
only domain    &   .388         &   .563        &   \textbf{.637}  \Bstrut \\
\hline
\end{tabular}%
}
\end{table}

\vspace{-0.1cm}
\subsection{Extrinsic Evaluation: Sexism Detection}

We next implement extrinsic evaluation to adjust our specialised \textsc{S}ex\textsc{We}s to a downstream binary classification task -- sexism detection -- which assesses the effectiveness of word embeddings with domain information.

\subsubsection{Dataset}

We use the only sexism dataset in Chinese, Sina Weibo Sexism Review (\textsc{Swsr}) \citep{jiang2022swsr}, with posts labeled for sexism from the Sina Weibo platform. \textsc{Swsr} annotations are constructed at different levels of granularity, and we use the binary labels: sexist and non-sexist. We split the entire dataset into training and test sets in the ratio of 4 to 1. We further randomly select 20\% of the training set as the validation set for model fine-tuning process, and finally utilise the whole training set to evaluate model capacity on test set. More details are shown in Table \ref{tab:dataset}. 

\begin{table}[h]
\centering
\caption{Distribution of train, validation and test sets, sexist text rate (\textsc{Str}) in the \textsc{Swsr} dataset.}
\label{tab:dataset}
\resizebox{0.33\textwidth}{!}{%
\begin{tabular}{lccc}
\hline
\textbf{} & \textbf{Train} & \textbf{Validation} & \textbf{Test} \Tstrut\Bstrut \\ \hline
Sexist             & 2244           & 561                 & 288       \Tstrut    \\ 
Non-Sexist        & 4214           & 1053                & 609           \\ 
Total               & 6458           & 1614                & 897        \Bstrut   \\ \hline
\textsc{Str} (\%)                & 34.7           & 34.8                & 32.1           \Tstrut\Bstrut \\ \hline
\end{tabular}%
}
\end{table}

\vspace{-0.1cm}
\subsubsection{Sexism Detection Models Tested}

We leverage a simple text-based Convolutional Neural Network (\textsc{Tcnn}) \citep{kim2014convolutional} as our primary classifier, which is a popular architecture for dealing with NLP tasks with a good feature extraction capability \citep{zhang2021research}, leading to a smaller number of parameters, lower computational needs, and a faster training speed \citep{zhang2021research}. \textsc{Tcnn} is fed with different vectors used in the intrinsic evaluation, or changed to other state-of-the-art models for comparison to demonstrate the impact of our specialised embeddings on detecting sexist text. For vectors, we use the original and specialised word embeddings evaluated in the intrinsic experiments in combination with static \textsc{Bert} embeddings extracted from Chinese \textsc{Bert}\footnote{We extract contextualised \textsc{Bert} embeddings from the initial embedding layer of Chinese \textsc{Bert} trained on \textsc{Swsr} training set, using Huggingface \textsc{Bert} model `hfl/chinese-bert-wwm-ext'.}.

As baseline models, we use \textsc{Bert} \citep{devlin2019bert} and a state-of-the-art Chinese pre-trained model \textsc{Macbert}\footnote{https://huggingface.co/hfl/chinese-macbert-base}, which performs better than normal Chinese \textsc{Bert}\footnote{https://huggingface.co/bert-base-chinese} and other variants in some classification tasks \citep{cui2020revisiting}.

\subsubsection{Evaluation Setup}

We use the Adam optimiser (0.0001 learning rate) and a maximum sequence length of 100 for all baseline models. \textsc{Tcnn} contains 128 units in the hidden layer with the dropout value 0.4, and we use Huggingface models `bert-base-chinese' (\textsc{Bert}) and `hfl/chinese-macbert-base' (\textsc{Macbert}). We train the \textsc{Tcnn}-based models for 100 epochs and \textsc{Bert}-based models for 4 epochs, using the same batch size of 32. We report the accuracy and macro F1 scores as the evaluation metrics.

\subsubsection{Analysis of Results}

We report the results for sexism detection in Table \ref{tab:extrinsic}. We see that the classifier with our \textsc{S}ex\textsc{We}s achieves the highest F1 and accuracy scores, outperforming all baseline classifiers and classifiers with baseline retrofitted embeddings, and most of our models with different constraints display better results than baselines. The classifier with our \textsc{S}ex\textsc{We}s also exhibits stable performance with relatively small fluctuations in scores. 
Comparing baseline embeddings, there are notable improvements (0.093-0.135) in our \textsc{S}ex\textsc{We}s compared to those using \textsc{Fasttext} word embeddings and popular Chinese embeddings \textsc{Vcwe}, and better performance than \textsc{Bert} embeddings.
Additionally, our model slightly outperforms \textsc{Bert}-related models \textsc{Bert} and \textsc{Macbert}, but both of them present smaller fluctuations due to high stability. 
\textsc{Retrogan} shows the best results among all baseline specialisation models and outperforms all non-specialised embeddings, but it is still below our \textsc{S}ex\textsc{We}s. 
Moreover, we can draw some conclusions that are in line with the intrinsic evaluation. That is, leveraging sexism-related constraints and external constraints in target language for the cross-lingual specialisation process improves the detection of online sexism, and only using general constraints also shows the effectiveness in this task compared to other specialisation baselines.

\begin{table*}[h]
\centering
\caption{Results of sexism detection with standard deviations (average of 10 runs).}
\label{tab:extrinsic}
\resizebox{0.8\textwidth}{!}{%
\begin{tabular}{llcccc}
\hline
\textbf{Model} & \textbf{}         & \textbf{F1-sex} & \textbf{F1-not} & \textbf{Macro-F1} & \textbf{Accuracy} \Tstrut\Bstrut \\ \hline 
\multirow{6}{*}{Baseline embeddings} & +\textsc{Ft}               &  .483 (±.015)                      &  .723 (±.044)                      &  .603 (±.028)                        &  .641 (±.040)                       \Tstrut \\
& +\textsc{Vcwe}             &  .355 (±.149)                      &  .796 (±.010)                      &  .645 (±.071)                        &  .682 (±.008)                        \\
& +\textsc{Bert}\_emb        &  .573 (±.059)                      &  .835 (±.009)                      &  .704 (±.027)                        &  .765 (±.006)                        \\
& +\textsc{Ar}               &  .490 (±.025)                      &  .840 (±.017)                      &  .668 (±.009)                        &  .770 (±.011)                        \\
& +\textsc{Retrogan}          &  .622 (±.010)                      &  .811 (±.056)                      &  .717 (±.027)                        &  .753 (±.044)                        \\
& +\textsc{Clsri}             &  .638 (±.005)                      &  .775 (±.010)                      &  .707 (±.006)                        &  .723 (±.007)                     \Bstrut \\ \hline 
\multirow{2}{*}{Baseline models} & \textsc{Bert}              &  .641 (±.006)                      &  .782 (±.008)                      &  .711 (±.006)                        &  .729 (±.007)                       \Tstrut \\
& \textsc{Macbert}           &  \textbf{.658 (±.013)}             &  .789 (±.015)                      &  .724 (±.013)                        &  .739 (±.014)                    \Bstrut \\ \hline 
\multirow{4}{*}{\textsc{S}ex\textsc{We}s} & \multicolumn{1}{l}{\textbf{\textsc{S}ex\textsc{We}s}} &  .626 (±.035)                      &  \textbf{.849 (±.008)}             &  \textbf{.738 (±.016)}               &  \textbf{.786 (±.008)}              \Tstrut \\
& \multicolumn{1}{r}{w/o external}   &  .627 (±.041)                      &  .840 (±.044)                      &  \textbf{.738 (±.024)}               &  .761 (±.034)                        \\
& \multicolumn{1}{r}{only general}      &  .622 (±.061)                      &  \underline{.842 (±.011)}                &  .732 (±.030)                        &  \underline{.779 (±.012)}                  \\
& \multicolumn{1}{r}{only domain}       &  \underline{.646(±.011)}                 &  .817 (±.056)                      &  \underline{.733 (±.032)}                  &  .764 (±.046) \Bstrut \\  \hline   
\end{tabular}%
}
\end{table*}

\subsubsection{Impact of Class Imbalance}

Sexism or abuse tends to be the minority class in most datasets. In the case of the \textsc{Swsr} dataset, 65.5\% are non-sexist instances \citep{jiang2022swsr}. Our \textsc{S}ex\textsc{We}s F1 score for the sexist class is 0.626, which is still clearly below the F1-not score of 0.849. We can also clearly observe that the F1-scores between sexist and non-sexist classes differ greatly, with the average F1-score of the non-sexist class being about 0.227 higher than that of the sexist class.
This shows a negative impact of class imbalance on the sexism detection task, and the potential challenges that sexist texts may bring to the detection (see more in the subsection Qualitative Analysis).

Resampling and data augmentation techniques could be considered to mitigate the imbalance in the future \citep{bigoulaeva2022addressing,rizos2019augment}.


\subsubsection{Qualitative Analysis}

In addition to quantitative evaluation, we also conduct a qualitative analysis of some misclassified cases to assess the potential of \textsc{S}ex\textsc{We}s for sexism detection as well as the challenges.
When looking at predicted examples from \textsc{Bert} and classifiers with original embeddings and our \textsc{S}ex\textsc{We}s, we see some recurrent types of misclassification in Table \ref{tab:error}.

\begin{table*}[h]
\centering
\caption{Misclassified examples by three models: \textsc{Tcnn} + \textsc{Fasttext} embeddings (\textsc{Tcnn}+\textsc{Ft}), \textsc{Bert}, and \textsc{Tcnn} + specialised embeddings (\textsc{S}ex\textsc{We}s), along with ground truth labels.}
\label{tab:error}
\resizebox{\textwidth}{!}{%
\begin{tabular}{lcccc}
\hline
\textbf{Text}                                                                                                                                                                      & \textbf{\textsc{Tcnn}+\textsc{Ft}} & \textbf{\textsc{Bert}} & \textbf{\textsc{S}ex\textsc{We}s} & \textbf{Ground Truth} \Tstrut\Bstrut \\ \hline
\begin{tabular}[c]{@{}l@{}}1. 尊重驴不带套的权力，意外怀孕的权力，尊重就vans。\Tstrut \\ \texttt{Translation}: Respect the rights of women without wearing condoms and \\ unintended pregnancies, that's it. \Bstrut\end{tabular} & Non-Sexist          & Non-Sexist    & Sexist          & \textbf{Sexist}       \\ \hline
\begin{tabular}[c]{@{}l@{}}2. 学历高的估计更厉害，从道理上说服你，不然就身体上睡服你。 \Tstrut \\ \texttt{Translation}: Males with higher education may be better at persuading you \\ or f*cking you. \Bstrut \end{tabular}             & Non-Sexist          & Non-Sexist    & Non-Sexist      & \textbf{Sexist}       \\ \hline
\begin{tabular}[c]{@{}l@{}}3. 田园女权，女拳师，极端女权，是我是我都是我。\Tstrut \\ \texttt{Translation}: Pastoral feminist, female boxer, extreme feminist, it is all me. \Bstrut \end{tabular}                                 & Sexist              & Sexist        & Sexist          & \textbf{Non-Sexist}   \\ \hline
\end{tabular}%
}
\end{table*}


\textit{(i) Implicit sexism:} 
Humour, irony and sarcasm are difficult to be identified. Example (1) is sexist irony without an explicitly abusive expression. The model with \textsc{S}ex\textsc{We}s successfully deemed it sexist, while the others failed. 

\textit{(ii) Informal and code-mixed expressions:} 
Example (1) is a Chinese-English code-mixed text, and the slang word `vans' in English has a similar pronunciation as `完事了' (that's it) in Chinese. `驴' usually refers to `donkey', but it is commonly used in sexist expressions that offend women\footnote{`驴' comes from `婚驴' (marriage donkey), and is intended to depict the image of `women who are as stupid as donkeys in marriage, deprived of a lot of benefits, but still enjoy silly happiness’.}.

\textit{(iii) Implicit attack target:} 
The attack target might not explicit appear like example (2). All models failed to predict it as sexist text. It demeans the group of highly educated males, but the target can only be guessed from the context. 

\textit{(iv) Homophones:}
Homophones are common in sexist speech to convey abusive connotations, or to obfuscate and avoid detection. `说服' and `睡服' have the same pronunciation in (2). `说服' is a general term that means persuade or convince, and `睡服' is a homophonic word with a similar meaning to persuade someone by f*cking.

\textit{(v) Overuse of explicit sexist words:} 
Sexist words might be overused in one text, leading to the over-dependence of the model on these words, while sexist targets in posts are confounding and hard to be identified. All models failed in example (3), and we see that the model can easily deem a text sexist if it contains many sexist words, despite not having a specific targeted individual or group.

\section{Discussion}

\subsection{Visualisation of Word Embeddings}


\begin{figure*}[ht!]
  \centering
  \includegraphics[width=\linewidth]{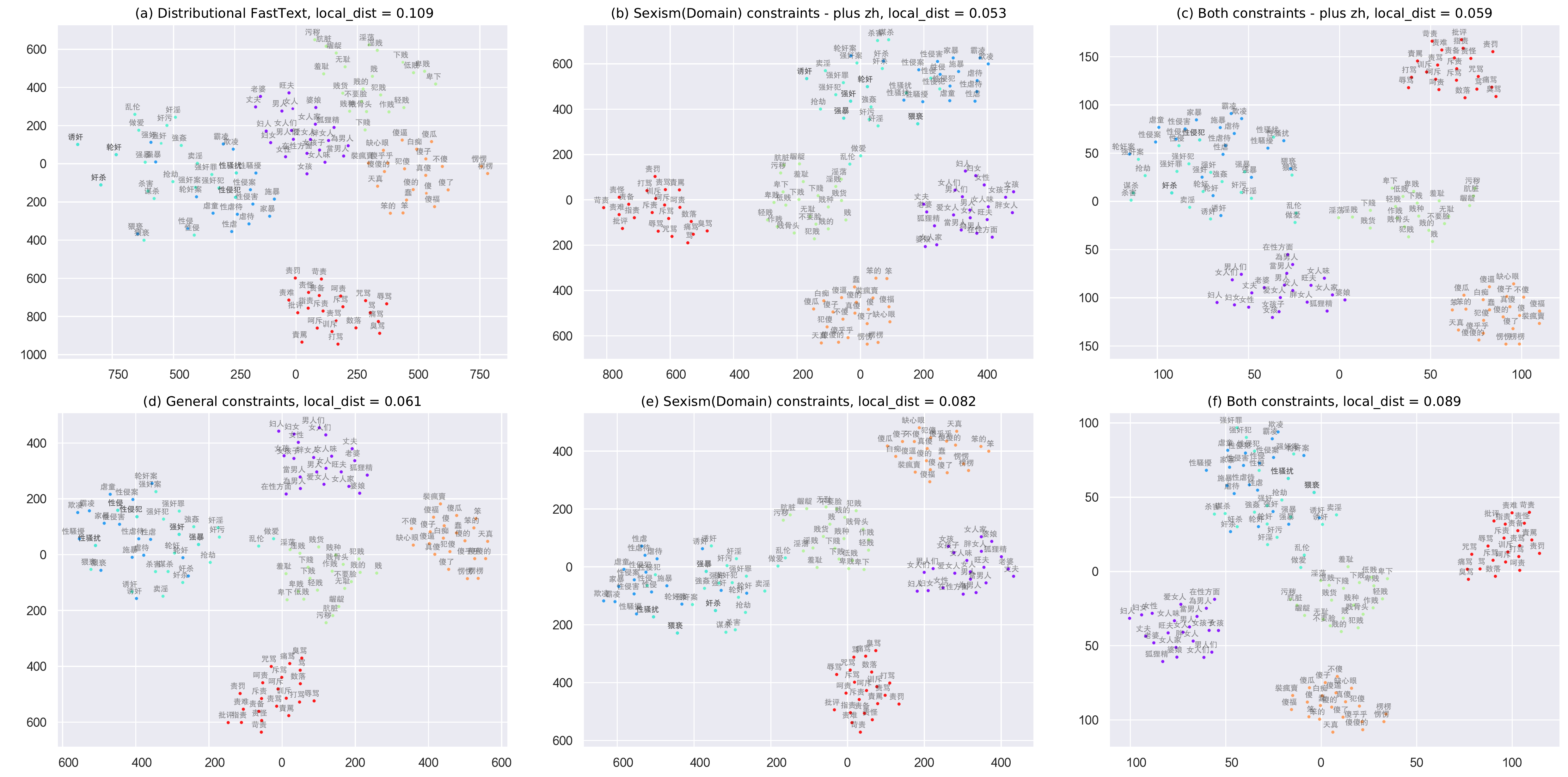}
  \caption{t-\textsc{Sne} visualisations of \textsc{S}ex\textsc{We}s word embeddings. Each color group indicates a Chinese domain word with its 20 neighbours generated from original \textsc{Fasttext} vectors. There are totally 6 seed words selected, namely purple for 女人 (woman), blue for 性侵 (sexual assault), skyblue for 强奸 (rape), green for 下贱 (b*tchy),  orange for 傻 (stupid), and red for 责骂 (scold). Averaged local distance of word clusters (local\_dist) is measured based on the t-\textsc{Sne} space.}
  \label{fig:emb.visualise}
\end{figure*}

We visualise both original \textsc{Fasttext} embeddings and various specialised \textsc{S}ex\textsc{We}s embeddings.
We select six sexism-related seed words and gather each seed word with its 20 nearest neighbors from the initial word vector space, to explore changes in these domain word groups during our specialisation process. Figure \ref{fig:emb.visualise} shows the visualisation of word embeddings with dimensional reduction by t-\textsc{Sne} \citep{laurens2008tsne}. 
To further investigate the semantic shift between different word vector spaces \citep{hamilton2016cultural}, we measure the average cosine distance between a seed word and its neighbours in each local word cluster, and average distances among the six clusters to obtain the overall distance in the space. The local distance is presented in subplot titles of Figure \ref{fig:emb.visualise}.

Looking at both the spatial range of visualised word clusters and local distances, we can observe that all specialised groups of domain words become more independent and get closer from the original distributional vector space in Figure \ref{fig:emb.visualise} (a) to any of our specialised vector space (see Figure \ref{fig:emb.visualise} (b)-(f)), which illustrates the benefit of our specialisation method. After the specialisation process with English constraints, the distance of word clusters shows a significant decrease, further decreasing after adding external Chinese constraints.
For word embeddings that incorporate more domain information (Figure \ref{fig:emb.visualise} (e) and (f)), the connections between words in each cluster become stronger, compared to embedding spaces that are only retrofitted with knowledge of general constraints in Figure \ref{fig:emb.visualise} (d). Furthermore, after adding external Chinese constraints, the vector space specialised only with domain knowledge becomes more contiguous (see Figure \ref{fig:emb.visualise} (e) to (b)), while the spaces specialised by both constraints are relatively sparse (see Figure \ref{fig:emb.visualise} (f) to (c)). This opposite change may be caused by perturbations of commonsense knowledge, since general constraints outnumber domain constraints.



\subsection{Ablation Study}

To evaluate different components, we perform a study of the following ablated models of \textsc{S}ex\textsc{We}s: \textit{(i) removing phrase-level projection; (ii) removing constraint refinement;} and \textit{(iii) removing \textsc{Retrogan} post-specialisation}.
In Table \ref{tab:ablation}, we can see that our model outperforms all ablated models, which demonstrates the important contribution of all components.
Although phrase-level constraint processing in the projection step does not significantly improve the quality of embeddings, this step validates the positive impact of doing domain-related phrase mapping on identifying sexism.
The results also highlight the effectiveness of \textsc{Stm} in refining the noisy lexico-semantic relations between constraints compared with the one without constraint refinement.
Furthermore, we can validate the capability of \textsc{Retrogan} post-specialisation step to efficiently apply the retrofitting mapping to full word vector space when comparing to specialised word vectors without post-specialisation step.


\begin{table}[h]
\centering
\caption{Results for \textsc{S}ex\textsc{We}s and ablative methods.}
\label{tab:ablation}
\resizebox{0.47\textwidth}{!}{%
\begin{tabular}{lccccc }
\hline
                   & \multicolumn{3}{c}{\textbf{Intrinsic}}                            & \multicolumn{2}{c}{\textbf{Extrinsic}} \Tstrut\Bstrut  \\ \cline{2-6} 
 & \textbf{SL999}                                                        & \textbf{WS240}                                                        & \textbf{WS296}                                               &   \textbf{Macro-F1}        &   \textbf{Acc.}          \Tstrut\Bstrut   \\ \hline
\textbf{\textsc{S}ex\textsc{We}s}    &   \textbf{.406}  &   \textbf{.586} &   .608  &   \textbf{.738}  &   \textbf{.786} \Tstrut \\ 
w/o phrase         & .404                                                         & .571                                                         & \textbf{.611}                                       &   .726           &   .778           \\ 
w/o refinement     &   .390           &   .536           &   .591  &   .713           &   .768           \\ 
w/o \textsc{Retrogan}        &   .398           &   .529           &   .594  &   .704           &   .760        \Bstrut \\ \hline
\end{tabular}%
}
\end{table}

\vspace{-0.1cm}

\subsection{Performance versus Complexity Trade-off Analysis}

According to experimental results, the overall performance of \textsc{S}ex\textsc{We}s fine-tuned by our cross-lingual domain-aware specialisation system shows 0.004-0.065 correlation score improvement in word similarity benchmark and 0.014-0.135 F1 score improvement in sexism detection.
The results of both intrinsic and extrinsic evaluations demonstrate the effectiveness of specialised word vectors compared to pre-trained word vector baselines, and show improved performance over all other specialisation systems with similar model complexity.
Compared with \textsc{Bert}-related baselines, our \textsc{S}ex\textsc{We}s is based on a simple \textsc{Tcnn} architecture and still achieves a slight increase in the performance of detecting sexist content, showing further potential for more advanced and robust networks.
Furthermore, we only need to train once to construct sexist word embeddings. Instead of only using it for sexism detection, it can also be reused to study sexism-related issues.
Only by collecting new constraints, the methodology of building cross-lingual specialisation system can be further transferred to other low-resourced domains to detect abnormal behaviours online.

\section{Conclusion and Future Work}

To tackle sexism detection for low-resource languages, we propose an effective system for cross-lingual domain-aware semantic specialisation by injecting external constraints referring to sexist terms in both source and target languages. We report notable performance of \textsc{S}ex\textsc{We}s in both intrinsic and extrinsic evaluations, visualising the positive trend of word embeddings during the specialisation, as well as through an ablation study. 
However, we only observe a modest improvement after adding cross-lingual constraints, potentially due to its limited size.
In the future, we plan to explore full automation of cross-lingual constraint creation
and the extension of our approach to contextualised embeddings.

\section{Ethical Considerations}

Online sexism and abuse are sensitive subjects with various ethical concerns in the controversy surrounding the freedom of speech. To develop the fairness and reliability of our work, we address the following limitations:

\begin{itemize}
    \item \textbf{Confidentiality:} Accessing the data is essential to make our work effective. Since the data is already public, to address the trade-off between privacy and effectiveness, original data has all personally identifiable data removed to ensure user anonymity.
    \item \textbf{Potential for harm:} Our work is not intended to harm vulnerable groups who are already discriminated against. While one could make bad use of sexism detection systems, such as learning to circumvent detection of their posts, our work is solely intended for the benign purposes of detection and mitigation of sexist speech.
    \item \textbf{Results communication:} Our work is free of plagiarism or research misconduct, but acknowledge potential limitations when analysing social media data, especially sexist data that does not clearly represent the attack target.
\end{itemize}

\section{Acknowledgements}

We would like to thank the anonymous reviewers for all their valuable comments and suggestions.
Aiqi Jiang is funded by China Scholarship Council (CSC Funding, No. 201908510140). The authors acknowledge the support of the CONICET–Royal Society International Exchange (IEC\textbackslash R2\textbackslash 192019). This research utilised Queen Mary’s Apocrita HPC facility, supported by QMUL Research-IT https://doi.org/10.5281/zenodo.438045.

\bibliography{custom}

\end{CJK*}
\end{document}